\documentclass[
]{ceurart}

\usepackage{listings}
\usepackage{float}
\usepackage{graphicx} \usepackage{tikz} \usetikzlibrary{ 
   positioning, 
   arrows.meta, 
   shapes.geometric, 
   calc 
} 
\usepackage{booktabs}

\begin{document}


\conference{  }

\title{Natural Language Programming:\\
Rethinking Language, Memory, and Cognition\\
in the Era of Generative AI}


\author[1]{Giovanni Adorni}[
orcid=0000-0003-3933-4377,
email=giovanni.adorni@unige.it,
]
\cormark[1]
\fnmark[1]
\address[1]{DIBRIS, University of Genoa, Viale F.Causa 13, 16145 Genoa, Italy}

\author[2]{Daniele Grosso}[
orcid=0000-0002-0699-3459,
email=daniele.grosso@unige.it,
]
\cormark[1]
\fnmark[1]
\address[2]{DIFI, University of Genoa, Via Dodecaneso 33, 16146 Genoa, Italy}

\cortext[1]{Corresponding author.}
\fntext[1]{These authors contributed equally.}

\begin{abstract}
Generative Artificial Intelligence is transforming the role of natural language in computational systems. Within the classical paradigm of Natural Language Processing (NLP), language has primarily been treated as an object of computation: something to analyse, classify, translate, or generate. The emergence of Large Language Models (LLMs) challenges this perspective by enabling natural language to function increasingly as an executable cognitive medium through which humans can structure reasoning, orchestrate problem solving, and regulate hybrid human--AI workflows.

This paper introduces the concept of \textit{Natural Language Programming (NLPg)} as a theoretical and epistemological extension of traditional NLP. Rather than focusing on language understanding alone, NLPg conceptualises prompts as dynamic cognitive artefacts that externalise goals, constraints, assumptions, validation criteria, and intermediate reasoning states. In this perspective, prompting is not reduced to a technical interaction technique, but becomes a form of dialogic cognitive programming operating within distributed human--AI memory architectures.

Building on theories of distributed cognition, extended mind, metacognition, and dialogic learning, the paper proposes the notion of Dialogic Working Memory, describing how prompts and conversational states function as temporary external memory structures supporting reasoning continuity and reflective regulation. We further map existing memory architectures (BTMA, SKGE, RMR) onto this framework, illustrating how episodic, semantic, and metacognitive layers support long-horizon conversational coherence.

Finally, implications for computational linguistics and AI-mediated knowledge systems are discussed, suggesting the emergence of a post-symbolic and post-interface paradigm in which natural language operates simultaneously as communication medium, executable specification, cognitive scaffold, and distributed memory architecture.
\end{abstract}


\maketitle

\section{Introduction}

For several decades, Natural Language Processing (NLP) has primarily conceptualised language as an object of computation: a symbolic or statistical entity to be parsed, classified, translated, summarised, or generated \cite{jurafsky2009speech,winograd1972understanding}. From symbolic systems to transformer architectures, the dominant paradigm has remained substantially stable: language is treated as information to be processed computationally.

Recent advances in Generative Artificial Intelligence (GenAI), particularly Large Language Models (LLMs), challenge this assumption. While traditional NLP systems focused on recognizing or generating linguistic structures, LLM-based systems increasingly enable natural language to function as an operational medium through which humans formulate intentions, coordinate reasoning, and regulate interaction with computational systems. Language is therefore no longer merely processed: it becomes a vehicle for orchestrating cognitive workflows.

This transformation suggests a broader epistemological shift within computational linguistics. Interaction with LLMs increasingly resembles a form of cognitive orchestration in which prompts externalise goals, constraints, intermediate reasoning states, and reflective strategies \cite{liu2022pretrain,wei2022,brown2020gpt3}. Natural language may therefore operate as a mechanism for dynamically configuring hybrid human--AI cognitive systems.
This perspective also resonates with emerging Cyber Humanities approaches, which interpret generative AI not merely as a technological tool, but as a cognitive and cultural infrastructure reshaping human agency, knowledge mediation, and reflective interaction \cite{adornibellini2025}.

The limitations of the classical NLP paradigm become particularly evident when considering memory and context. Traditional architectures largely focused on sentence-level processing and probabilistic prediction, often treating context as an auxiliary parameter. By contrast, interaction with LLMs increasingly relies on persistent conversational states, iterative refinement, contextual reconstruction, and dialogic continuity, progressively acquiring characteristics associated with working memory and cognitive architectures.

Theoretical perspectives from distributed cognition and philosophy of mind provide important tools for understanding this transition. Distributed cognition argues that cognitive processes emerge through interactions among agents, artefacts, and environments \cite{hutchins1995}, while the "Extended Mind" thesis suggests that external artefacts may become integral components of cognition itself \cite{clark1998extended}. Within this framework, prompts, conversational traces, and interaction histories can be interpreted as forms of externalised working memory distributed across hybrid human--AI systems.

At the same time, dialogic and socio-cultural perspectives emphasize that meaning emerges through situated interaction and co-construction processes rather than isolated symbolic manipulation \cite{bakhtin1981,vygotsky2012thought}. Prompting therefore increasingly resembles a dialogic process through which reasoning is stabilised, negotiated, and reconstructed across conversational exchanges.

Building on these premises, this paper introduces the concept of \emph{Natural Language Programming} (NLPg) as a theoretical extension of traditional NLP. Rather than treating language exclusively as an object of computation, NLPg conceptualizes prompts as executable cognitive artefacts through which humans externalise reasoning, maintain contextual continuity, and orchestrate hybrid cognitive workflows. 
More specifically, we argue that prompts simultaneously operate as:
\begin{itemize}
    \item executable specifications,
    \item cognitive artefacts,
    \item metacognitive scaffolds,
    \item and dialogic memory structures.
\end{itemize}
To formalize this perspective, the paper proposes the notion of \emph{Dialogic Working Memory}, describing how conversational interaction with LLMs supports the stabilization and reconstruction of distributed cognitive states.
\paragraph{Positioning within existing research -} Recent research on prompting and conversational AI has primarily focused on prompt optimization, chain-of-thought reasoning, code generation, and interaction efficiency \cite{wei2022,reeves2025prompts}. Within this landscape, the term Natural Language Programming has often referred to the use of natural language for specifying computational tasks or generating executable code.

Our proposal extends this perspective by interpreting NLPg as a distributed cognitive and dialogic memory architecture operating through conversational interaction rather than as a simple code-generation paradigm.


Existing approaches such as Retrieval-Augmented Generation [12], In-Context Learning [5], and tool use [13] provide mechanisms for external memory, dynamic adaptation, and action coordination. Dialogic Working Memory can be interpreted as a higher-level orchestration layer that integrates and regulates these mechanisms within a conversational cognitive framework.

Unlike existing work primarily focused on prompt optimization or code generation, this paper conceptualizes NLPg as a distributed cognitive infrastructure in which prompts externalise reasoning processes, stabilize contextual continuity, and support reflective orchestration across hybrid human--AI systems.

\paragraph{Contributions -}
The main contributions of this paper are:
\begin{enumerate}
    \item A theoretical definition of \textbf{NLPg} as an extension of NLP toward language-mediated cognitive orchestration.
    \item The proposal of \textbf{Dialogic Working Memory}, interpreting conversational interaction as a form of reconstructive distributed memory management.
    \item A conceptual mapping of conversational memory architectures (BTMA, SKGE, RMR) onto the NLPg framework.
    \item A discussion of implications for computational linguistics, suggesting a transition toward cognitive mediation infrastructures rather than purely text-processing systems.
\end{enumerate}

The remainder of the paper is organised as follows. Section~2 discusses the transition from NLP to NLPg. Section~3 introduces prompting as cognitive programming. Section~4 presents the Dialogic Working Memory framework. Section~5 examines NLPg as a metacognitive scaffold. Section~6 discusses implications for computational linguistics, and Section~7 concludes with limitations and future directions.

\section{From NLP to NLPg: Processing vs. Orchestration}

The emergence of LLMs is transforming not only the technological
landscape of NLP, but also the assumptions underlying the relationship
between language, cognition, and computation. While traditional NLP
primarily treated language as an object of computational analysis,
interaction with generative models increasingly positions natural
language as a medium for orchestrating cognitive processes and
configuring hybrid human--AI workflows.

Classical NLP systems were designed to process linguistic structures
through parsing, classification, translation, and information
extraction \cite{jurafsky2009speech}. Even with contextual embeddings
and transformer architectures, language largely remained input data to
be statistically modeled and generated. By contrast, interaction with
LLMs relies on natural language to define objectives, structure
reasoning paths, maintain contextual continuity, and regulate
iterative problem solving. Through prompting, users employ language
not simply to request information, but to coordinate reasoning across
conversational exchanges.

This shift can be interpreted as a transition from
\emph{language processing} to \emph{language orchestration}. In the
former paradigm, computational systems aim to understand or generate
linguistic content. In the latter, natural language functions as an
operational medium for coordinating distributed cognitive processes
across human and artificial agents.

\begin{table}[!ht]
\centering
\caption{From Natural Language Processing to Natural Language Programming}
\label{tab:nlp-nlpg}
\begin{tabular}{p{0.43\linewidth}p{0.43\linewidth}}
\toprule
\textbf{Natural Language Processing (NLP)} & \textbf{Natural Language Programming (NLPg)} \\
\midrule
Language as object of computation & Language as executable cognitive medium \\
Text as input data & Prompt as cognitive artefact \\
Context as processing parameter & Context as conversational state \\
Output generation & Iterative cognitive regulation \\
Evaluation of final output & Traceability of reasoning process \\
\bottomrule
\end{tabular}
\end{table}

Table~\ref{tab:nlp-nlpg} summarizes the conceptual differences between traditional NLP and the proposed NLPg paradigm. The distinction is not merely terminological. Traditional programming languages rely on
explicit syntax and deterministic execution rules, whereas prompting operates through flexible and context-dependent linguistic structures.
Despite their apparent informality, prompts increasingly behave as executable specifications capable of configuring complex behaviors, including decomposition strategies, evaluation criteria, and iterative
refinement processes \cite{reeves2025prompts}.

This perspective resonates with Winograd and Flores
\cite{winograd1987understanding}, who argued that language should be
understood not merely as representation, but as situated action
embedded in social coordination. Contemporary LLM-mediated interaction
partially realizes this intuition: language increasingly operates as a
mechanism for coordinating distributed cognitive ecologies.

From this standpoint, prompting can be interpreted as a form of
\emph{Natural Language Programming}, where natural language assumes
functions traditionally associated with programming and cognitive
coordination. NLPg does not replace traditional programming paradigms;
rather, it introduces a complementary layer in which prompts stabilize
conversational states, orchestrate reasoning trajectories, and
externalize distributed cognitive processes.

This transformation also modifies the role of computational systems
themselves. LLMs increasingly behave not simply as language processors,
but as conversational cognitive infrastructures supporting contextual
reconstruction, reflective reasoning, distributed memory management,
and metacognitive regulation. The transition from NLP to NLPg therefore
marks not only a technological evolution, but a broader conceptual
shift from language as representational medium to language as
operational cognitive infrastructure.

\section{Prompting as Cognitive Programming}

The widespread adoption of LLMs has transformed prompting from a
marginal interaction mechanism into a central cognitive practice.
Prompts are no longer simple textual queries, but structured artefacts
through which users externalise intentions, organise reasoning,
regulate context, and coordinate iterative problem solving.

Prompting should therefore not be interpreted merely as query
optimization or interaction technique. Rather, it increasingly
resembles a form of \emph{cognitive programming}, in which natural
language configures reasoning spaces and coordinates distributed
cognitive processes across conversational interaction.

Unlike traditional programming languages, which rely on rigid syntax
and deterministic execution, prompting operates through flexible and
context-sensitive interaction. Yet prompts increasingly exhibit
functional characteristics analogous to programming abstractions,
including task decomposition, goal specification, constraint
definition, validation procedures, and iterative refinement.

This becomes particularly evident in advanced prompting strategies such
as chain-of-thought, reflective prompting, and multi-step workflows.
In these contexts, prompts do not simply request information; they
structure the cognitive trajectory through which reasoning is
generated, evaluated, and progressively refined.

\paragraph{Prompts as Cognitive Artefacts -}
The notion of \emph{cognitive artefact} \cite{norman1993things} provides a
theoretical foundation. Norman defines cognitive artefacts as
artificial structures designed to maintain, display, or manipulate
information in support of cognitive activity. Distributed cognition
theories further argue that cognition emerges through interactions
among agents, tools, and representations rather than remaining confined
within the individual mind \cite{hutchins1995}.

From this perspective, prompts can be interpreted as externalised
cognitive structures that temporarily stabilize goals, assumptions,
constraints, reasoning procedures, evaluation criteria, and reflective
strategies. Consequently, prompts operate not as isolated commands but
as transient representations of distributed cognitive states.

This interpretation introduces a key distinction between prompts as
\emph{instructions} and prompts as \emph{executable intentions}. In
traditional HCI, user inputs trigger predefined procedures. In
LLM-mediated environments, prompts often express partially
underspecified intentions that are progressively refined through
dialogic interaction. Execution becomes contextual, interpretative, and
conversationally negotiated.

\paragraph{Prompting and Metacognitive Regulation - }

Recent work on Prompt-Based Learning (PBL) and Natural Language Programming suggests that prompting supports reflective problem decomposition and metacognitive regulation rather than simple output generation \cite{adorniscalatorsani2026,reeves2025prompts}. Prompt sequences often reveal iterative cycles of planning, monitoring, evaluation, debugging, and revision, closely resembling the self-regulatory mechanisms described in metacognitive learning theories \cite{panadero2017,schraw1994}.

Thus, prompting progressively evolves from interaction mechanism to
dialogic cognitive regulation process. Conversational traces support
the stabilization and reconstruction of reasoning continuity across
iterative exchanges, enabling forms of externalised reflective
reasoning distributed across human and computational agents.

Within this hybrid ecology, LLMs function not merely as language
processors but as conversational cognitive infrastructures
participating in distributed reasoning. Prompting acts as the
coordination layer through which humans and computational systems
jointly construct, maintain, and regulate evolving cognitive states.
This shift provides the conceptual foundation for the next section,
which introduces Dialogic Working Memory.

\section{Memory Architectures}

The transition from NLP to NLPg requires a corresponding rethinking of memory. 
In classical NLP systems, memory is generally treated either as a static repository of linguistic representations or as a transient computational buffer supporting inference and generation. By contrast, interaction with Large Language Models increasingly exhibits characteristics of distributed, reconstructive, and dialogically mediated memory processes.

In NLPg, language no longer functions exclusively as an object of
computation. Instead, it progressively operates as a mechanism for
stabilizing contextual continuity across conversational interaction.
Prompts, summaries, retrieved traces, and conversational histories
collectively behave as distributed memory structures supporting
reasoning continuity across time.

This perspective integrates theories of working memory, distributed
cognition, and dialogic reconstruction with recent prompt-based memory
architectures for LLMs
\cite{baddeley1986working,hutchins1995,clark1998extended,Grosso2026btma,Grosso2026rmr}.

\paragraph{Working Memory and Conversational State -}
The concept of working memory traditionally refers to the temporary
maintenance and manipulation of information required for reasoning and
cognitive activity \cite{baddeley1986working}. Similarly, LLM-based systems
maintain transient conversational states containing instructions,
constraints, semantic references, and contextual traces that guide
subsequent generations.

Unlike biological working memory, however, conversational state in LLMs is not internally persistent. 
It emerges dynamically through 
1) contextual token accumulation, 
2) prompt reconstruction,
3) retrieval mechanisms, 
4) external memory injection, and
5) iterative conversational reinforcement.

Conversational interaction with LLMs therefore increasingly behaves as
a form of \emph{externalised working memory management}, where prompts,
summaries, and contextual traces collectively stabilize transient
cognitive states across interaction cycles.

Recent prompt-based memory architectures explicitly formalize this
mechanism. The Biomorphic Temporal Memory Architecture (BTMA) proposed
in \cite{Grosso2026btma} introduces a hierarchical conversational memory model inspired
by biological memory consolidation systems. 
The architecture distributes conversational memory across episodic,
semantic, and metacognitive layers supporting different forms of
contextual reconstruction and cognitive continuity, as summarised in
Table~\ref{tab:memory-triad}.

\begin{table}[!ht]
\centering
\caption{Dialogic memory layers in NLPg architectures}
\label{tab:memory-triad}
\begin{tabular}{p{0.14\linewidth}p{0.12\linewidth}p{0.40\linewidth}p{0.18\linewidth}}
\toprule
\textbf{Layer} & \textbf{Component} & \textbf{Function in NLPg} & \textbf{Cognitive analogy} \\
\midrule
Episodic & BTMA & Temporal traces of interaction, decisions, constraints and conversational states & Episodic memory \\
Semantic & SKGE & Typed relations among entities, concepts, artefacts and projects & Semantic memory \\
Metacognitive & RMR & Consistency monitoring and reflective re-consolidation & Executive control \\
\bottomrule
\end{tabular}
\end{table}

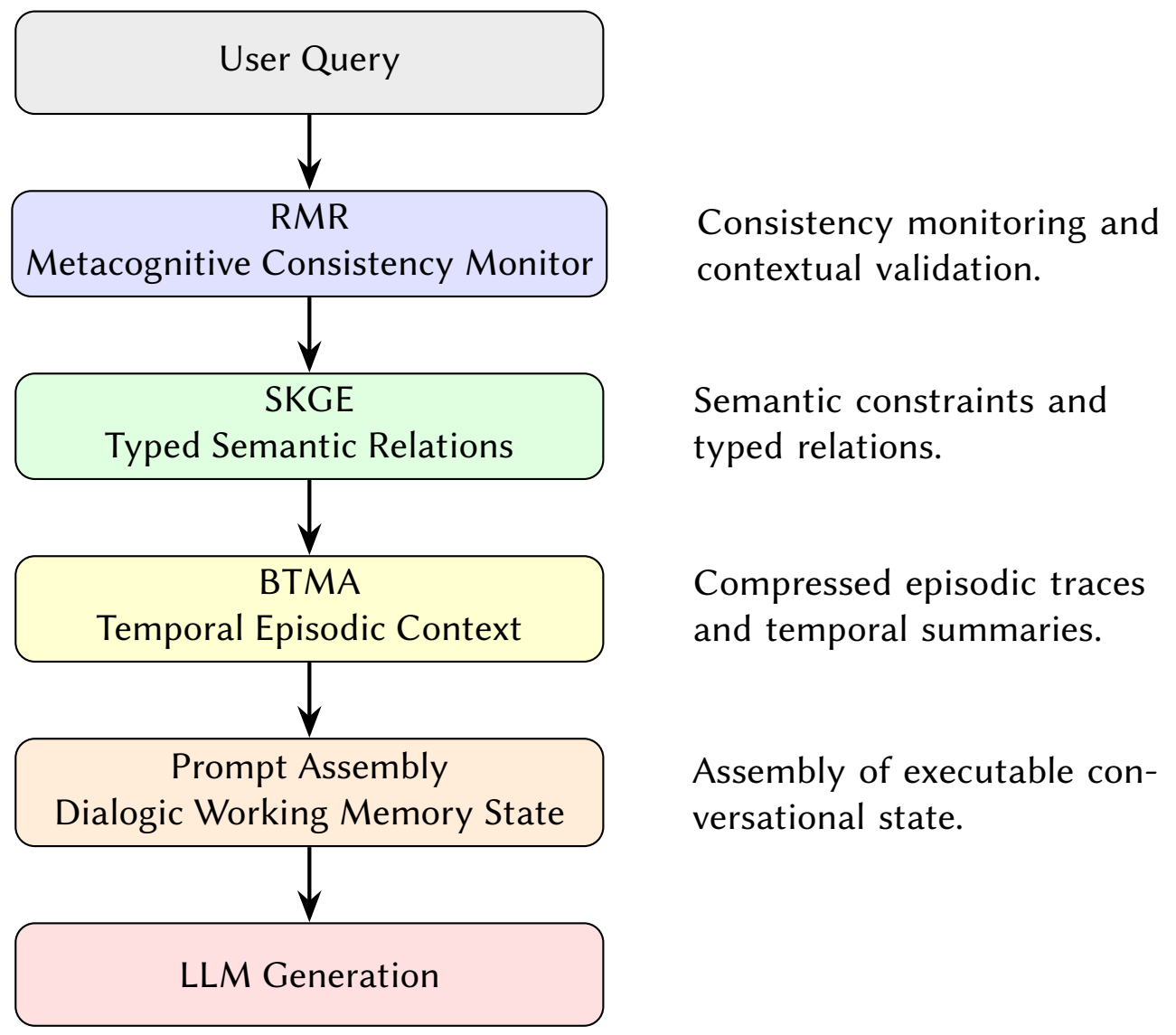
\begin{figure}[t]
\centering
\scriptsize
\begin{tikzpicture}[
    node distance=0.55cm,
    box/.style={
        rectangle,
        draw,
        rounded corners,
        minimum width=4.3cm,
        minimum height=0.75cm,
        align=center
    },
    note/.style={
        text width=3.5cm,
        font=\scriptsize,
        align=left
    },
    arrow/.style={-Stealth, thick}
]

\node[box, fill=gray!15] (query) {User Query};

\node[box, fill=blue!12, below=of query] (rmr)
{RMR\\ Metacognitive Consistency Monitor};

\node[box, fill=green!12, below=of rmr] (skge)
{SKGE\\ Typed Semantic Relations};

\node[box, fill=yellow!18, below=of skge] (btma)
{BTMA\\ Temporal Episodic Context};

\node[box, fill=orange!15, below=of btma] (prompt)
{Prompt Assembly\\ Dialogic Working Memory State};

\node[box, fill=red!12, below=of prompt] (llm)
{LLM Generation};

\draw[arrow] (query) -- (rmr);
\draw[arrow] (rmr) -- (skge);
\draw[arrow] (skge) -- (btma);
\draw[arrow] (btma) -- (prompt);
\draw[arrow] (prompt) -- (llm);

\node[note, right=0.55cm of rmr]
{Consistency monitoring and contextual validation.};

\node[note, right=0.55cm of skge]
{Semantic constraints and typed relations.};

\node[note, right=0.55cm of btma]
{Compressed episodic traces and temporal summaries.};

\node[note, right=0.55cm of prompt]
{Assembly of executable conversational state.};

\end{tikzpicture}
\caption{Dialogic memory injection pipeline for NLPg. Episodic, semantic, and metacognitive layers are assembled into a conversational state supporting long-horizon cognitive continuity before generation, supporting long-horizon conversational coherence and contextual reconstruction before generation.}
\label{fig:dialogic-memory-pipeline}
\end{figure}

As illustrated in Figure~\ref{fig:dialogic-memory-pipeline}, the BTMA/SKGE/RMR framework can be interpreted as a dialogic memory injection pipeline for NLPg. Episodic traces provide temporal continuity, semantic structures constrain interpretation, and metacognitive monitoring stabilizes contextual coherence before generation.

Within this framework, conversational traces are progressively
compressed and reorganised through hierarchical temporal structures,
including event-level traces, daily summaries, monthly consolidations,
and higher-level semantic abstractions.

The architecture mirrors the transition from working memory to
long-term consolidation in biological systems, where persistence is
achieved not through exhaustive retention but through selective
compression, contextual abstraction, and iterative re-consolidation.

A key implication of these architectures is that memory in NLPg is not
equivalent to storage. Conversational continuity emerges through the
reactivation and reconstruction of compressed linguistic traces rather
than through permanent persistence of raw interaction history.

\paragraph{Distributed and Dialogic Memory -}

The emergence of NLPg strongly resonates with theories of distributed
cognition \cite{hutchins1995} and the extended mind hypothesis
\cite{clark1998extended}. Within these frameworks, cognition is not confined
to internal mental processes, but distributed across external artefacts, symbolic representations, technological systems, and social interaction.

Prompting practices provide a particularly clear example of this
phenomenon. Prompts increasingly function as cognitive scaffolds that
externalise goals, constraints, intermediate reasoning states,
evaluation criteria, and metacognitive regulation strategies. In this
sense, prompts behave similarly to external memory artefacts such as
notes, conceptual maps, or symbolic inscriptions, reducing cognitive
load by stabilizing partial reasoning states outside the biological
brain.

Conversational interaction with LLMs also exhibits reconstructive
dynamics analogous to biological memory reconsolidation
\cite{nader2000,Lane2015MemoryRE}. Contextual information is not simply
retrieved unchanged. Instead, conversational histories are selectively
summarised, reinterpreted, compressed, and reintegrated into subsequent
interactions.

This process can be interpreted as a form of \emph{dialogic memory}:
a distributed memory mechanism emerging through iterative conversational
reconstruction. Memory continuity depends less on exact persistence
than on conversational coherence and contextual reactivation.

The Re-consolidative Metacognitive Architecture (RMR) explicitly
operationalizes this principle by introducing reflective procedures for
re-evaluating episodic and semantic traces across interaction cycles
\cite{Grosso2026rmr}. Within this framework, memory becomes selective,
interpretative, recursive, and dynamically reconstructed through
dialogic interaction.

From the perspective of NLPg, conversational memory architectures are
therefore significant not merely as persistence mechanisms, but as
cognitive mediation infrastructures supporting:

\begin{itemize}
    \item contextual continuity,
    \item distributed reasoning,
    \item reflective regulation,
    \item and long-horizon conversational coherence.
\end{itemize}

This evolution suggests that future computational linguistic systems
may increasingly focus not only on language understanding and
generation, but on the design of conversational cognitive
infrastructures capable of sustaining distributed reasoning processes across hybrid human--AI systems.

\section{NLPg as Metacognitive Scaffold}

Natural Language Programming (NLPg) redefines the relationship between
language, cognition, and learning. Prompting practices increasingly
function as metacognitive scaffolds supporting planning, monitoring,
reflection, and adaptive revision during problem-solving activities.
Unlike traditional interfaces designed for command execution,
LLM-mediated interaction supports iterative reasoning in which language
itself becomes a medium for organising thought. Prompts externalise
reasoning strategies and preserve intermediate conceptual states across
interaction cycles.

Prompt-Based Learning (PBL) frames prompting as a reflective process
through which learners iteratively construct, evaluate, and refine
reasoning strategies \cite{adorniscalatorsani2026,reeves2025prompts}.
Within this perspective, prompts function simultaneously as cognitive
organisers, external reasoning traces, metacognitive checkpoints, and
dialogic learning artefacts. The educational relevance lies in the
visibility of reasoning processes. Unlike traditional systems, NLPg
externalises substantial portions of reasoning through conversational
traces, supporting metacognitive awareness by enabling learners to
inspect strategies, revise assumptions, and identify inconsistencies.
Prompting therefore becomes not only a means of obtaining outputs, but
also a mechanism for organising reflective learning.

A key consequence of NLPg is the progressive traceability of reasoning
processes. Conversational traces document prompt evolution,
intermediate reasoning steps, decomposition strategies, revisions,
validation attempts, and reflective corrections. These traces
constitute \emph{epistemic artefacts}: externalised representations of
evolving cognitive processes. Metacognitive regulation includes
planning, monitoring, evaluation, and adaptive revision
\cite{schraw1994,panadero2017}. NLPg environments support these
mechanisms through recursive conversational workflows involving problem
formulation, decomposition, generation, validation, revision, and
re-generation. This transforms prompting into a form of dialogic
reflective regulation in which reasoning is progressively
co-constructed.

Within this framework, prompts behave similarly to metacognitive
operators: they activate reasoning procedures, constrain solution
spaces, establish evaluation criteria, and regulate conversational
direction. Reflective regulation is distributed across both human and
computational agents. The LLM contributes generative capabilities,
while the human performs epistemic supervision and contextual
validation. Traceability also introduces cognitive accountability.
Since reasoning remains partially externalised, users are encouraged to justify decisions and critically evaluate outputs rather than passively accepting them. This becomes particularly significant in the presence of hallucinations, semantic drift, and contextual inconsistencies in generative AI \cite{Grosso2026cdos}.
Conversational traceability supports validation by making reasoning trajectories inspectable and revisable.

Finally, NLPg enables the externalisation of metacognitive activity itself. Prompts, revisions, clarifications, and validation cycles form visible representations of cognitive regulation processes supporting reflective learning, collaborative reasoning, process-oriented assessment, and AI literacy development. However, this transformation also introduces risks. Externalisation may encourage over-reliance on conversational scaffolds, excessive delegation of reasoning, or superficial validation practices. Consequently, NLPg requires forms of reflective AI literacy capable of preserving human epistemic agency and critical cognitive autonomy within AI-mediated environments. 
This concern aligns with Cyber Humanities perspectives emphasizing the need to preserve reflective human agency and critical autonomy within AI-mediated cognitive ecosystems \cite{adornibellini2025}.
The central problem is no longer whether AI systems generate correct outputs, but whether users maintain the capacity to critically regulate and validate distributed reasoning processes. NLPg therefore repositions language from a medium for information exchange to a mechanism for orchestrating metacognitive activity across hybrid human--AI systems.

\section{Implications for Computational Linguistics}

The emergence of NLPg challenges several foundational assumptions of computational linguistics. If natural
language increasingly functions not only as an object of computation but also as an operational cognitive medium, then the role of language technologies must be reconsidered.

For decades, computational linguistics has primarily focused on
language understanding, generation, semantic representation, and
syntactic parsing. Even transformer architectures largely remained within this paradigm \cite{jurafsky2009speech}. By contrast, LLM-mediated interaction suggests a different perspective in which natural language operates not simply as representational input, but as a mechanism for orchestrating reasoning, maintaining contextual continuity, coordinating human--AI workflows, and regulating reflective interaction.

This transformation shifts computational linguistics toward
\emph{cognitive mediation infrastructures}: systems that do not merely process linguistic content, but actively participate in the stabilization and reconstruction of distributed cognitive processes.

\paragraph{The Conversational Turn -}

The rise of LLM-based systems introduces a \emph{conversational turn}
within computational linguistics. Historically, NLP has primarily
focused on sentence- or document-level processing, while dialogue
systems often treated conversation as sequential information exchange.
By contrast, NLPg frames conversation as a cognitive process, memory
mechanism, reflective structure, and coordination protocol.
Conversational interaction supports contextual stabilization,
iterative reinterpretation, semantic negotiation, and collaborative
reasoning.

This perspective explains the growing relevance of long-horizon
consistency, memory persistence, referential stability, behavioral
coherence, and metacognitive traceability. These phenomena cannot be
fully understood through traditional text-processing paradigms because
they emerge at the level of distributed interaction.

Recent work on conversational memory architectures and CDOS phenomena suggests that contextual degradation may affect not only output quality, but cognitive continuity itself
\cite{Grosso2026cdos}. Semantic drift, referential
instability, and behavioral inconsistency increasingly represent
disruptions in cognitive mediation rather than simple generation
errors.

\paragraph{Toward Cognitive Mediation Infrastructures -}
The implications of NLPg extend beyond technical architecture toward a
broader epistemological transition. Classical symbolic AI relied on
explicit formal representations and deterministic execution rules,
while contemporary NLP largely replaced symbolic reasoning with
probabilistic modeling. Yet both paradigms preserved a distinction
between user, interface, command, and computational execution.

NLPg partially dissolves these distinctions. In conversational systems,
language acts simultaneously as representation and execution; prompts
function as communication and cognitive regulation; conversational
traces operate as both interaction history and distributed memory.
Natural language increasingly becomes the operational substrate through
which cognition is coordinated.

This transformation does not imply the disappearance of formal
computation. Rather, computational processes become progressively
embedded within conversational and context-dependent cognitive
ecologies mediated by natural language. Language technologies are no
longer interpreted solely as systems for understanding or generating
text, but as infrastructures for coordinating cognition, stabilizing
memory, supporting distributed agency, externalizing reasoning, and
mediating reflective interaction across hybrid human--AI systems.

NLPg therefore represents not simply a new interaction paradigm, but a
broader reconfiguration of the relationship between language,
cognition, memory, and computation. Traditional NLP tasks such as
parsing, semantic modeling, information extraction, and generation
remain fundamental, but increasingly operate within conversational
ecologies in which natural language functions not only as
representational content, but also as a mechanism for cognitive
coordination and contextual orchestration.

\section{Conclusion}

The emergence of Large Language Models is transforming the role of
natural language within computational systems. Throughout the history
of NLP, language has primarily been interpreted as an object of
computation. This paper argued that generative AI introduces a broader
epistemological shift in which natural language increasingly functions
as an executable cognitive medium supporting reasoning, memory
externalization, reflective regulation, and hybrid human--AI interaction.

Building on this transformation, we introduced the concept of
\emph{Natural Language Programming} (NLPg) as a theoretical extension
of traditional NLP. Unlike classical language-processing paradigms,
NLPg conceptualizes prompts and conversational interaction as
operational cognitive structures capable of stabilizing contextual
continuity, organizing distributed memory, and supporting
metacognitive regulation. Within this framework, prompts operate as
executable intentions, cognitive artefacts, dialogic memory traces,
and reflective scaffolds.

We further proposed the notion of \emph{Dialogic Working Memory} to
describe how conversational interaction with LLMs externalises
temporary cognitive states across distributed human--AI systems.
Integrating perspectives from distributed cognition, working memory
theory, and dialogic interaction, the framework interprets
conversational exchange as a process of contextual reconstruction and
cognitive orchestration.

Recent architectures such as BTMA, SKGE, and RMR suggest that
LLM-mediated systems are already evolving toward structured forms of
distributed memory management and metacognitive stabilization
\cite{Grosso2026btma,Grosso2026rmr}. At the same time, phenomena such
as Cost-Driven Output Substitution (CDOS) highlight how conversational
continuity may degrade under increasing contextual load
\cite{Grosso2026cdos}.

From this perspective, computational linguistics can no longer be
interpreted exclusively as the study of language understanding or text
generation. Language technologies increasingly function as cognitive
mediation infrastructures supporting distributed reasoning, memory
reconstruction, and reflective coordination across hybrid human--AI
systems. This transformation suggests the gradual emergence of a
post-symbolic and post-interface paradigm in which the distinction
between language, memory, interface, and execution progressively
blurs.

\paragraph{Limitations and Future Work -}
The proposed framework remains primarily theoretical and conceptual, and should be interpreted as an epistemological model intended to support future empirical and architectural research. Although primarily conceptual, the framework emerged from ongoing experimental work on conversational memory architectures and AI-mediated educational environments.
Dialogic Working Memory and conversational orchestration therefore require further formal modeling and experimental validation.

Future research may explore:
(i) conversational memory persistence and long-horizon consistency;
(ii) metacognitive traceability in human--AI interaction;
(iii) educational applications of NLPg within reflective learning
environments; and
(iv) conversational infrastructures capable of preserving human
epistemic agency while leveraging LLM capabilities.

More broadly, the transition from Natural Language Processing to
Natural Language Programming may represent not simply a technological
evolution, but a deeper transformation in how language itself is
understood within the relationship between humans, cognition, and
computation.
Within this perspective, NLPg may also contribute to the broader Cyber Humanities agenda aimed at rethinking human-centered cognition, education, and cultural mediation in the age of generative AI.

\section*{Declaration on Generative AI}
  
 \noindent
 During the preparation of this work, the authors used ChatGPT (GPT-5.5) in order to:  Paraphrase and reword, Improve writing style, Grammar and spelling check. After using this tool, the authors reviewed and edited the content as needed and take full responsibility for the publication’s content.

\bibliography{references}

@inproceedings{adornibellini2025,
  author    = {Adorni, G. and Bellini, E.},
  title     = {Towards a Manifesto for Cyber Humanities: Paradigms, Ethics, and Prospects},
  booktitle = {2025 IEEE International Conference on Cyber Humanities (IEEE-CH)},
  address   = {Florence, Italy},
  year      = {2025},
  pages     = {1--8},
  doi       = {10.1109/IEEE-CH65308.2025.11279659}
}

@inproceedings{adorniscalatorsani2026,
  author    = {Adorni, G. and Scala, D. and Torsani, S.},
  title     = {Programming Cognition with Language: Toward a Dialogic Architecture for AI-Mediated Learning},
  booktitle = {Procs of ITHET 2026 - 22nd Int. Conf. on Info. Technology Based Higher Education and Training},
  address   = {Inland Norway Univ.},
  year      = {2026},
  month     = {20-22 May},
}

@book{baddeley1986working,
  title={Working Memory},
  author={Baddeley, A.D.},
  isbn={9780198521334},
  lccn={86000846},
  series={Oxford psychology series},
  year={1986},
  publisher={Clarendon Press}
}

@book{bakhtin1981,
  author    = {Bakhtin, M.M.},
  title     = {The Dialogic Imagination: Four Essays},
  publisher = {Univ. of Texas Press},
  address   = {Austin, TX, USA},
  year      = {1981}
}

@article{brown2020gpt3,
  title={Language Models are Few-Shot Learners},
  author={Brown, Tom B and Mann, Benjamin and Ryder, Nick and others},
  journal={Advances in Neural Information Processing Systems},
  volume={33},
  pages={1877--1901},
  year={2020}
}

@article{clark1998extended,
  author    = {Andy Clark and David Chalmers},
  title     = {The Extended Mind},
  journal   = {Analysis},
  volume    = {58},
  number    = {1},
  pages     = {7--19},
  year      = {1998},
  publisher = {Oxford University Press},
  url       = {http://www.jstor.org/stable/3328150}
}

@misc{Grosso2026btma,
  author    = {Grosso, Daniele},
  title     = {{BTMA}: Biomorphic Temporal Memory Architecture for {LLM}s},
  year      = {2026},
  publisher = {Zenodo},
  doi       = {10.5281/zenodo.18928639},
  note      = {Preprint}
}

@misc{Grosso2026rmr,
  author    = {Grosso, Daniele},
  title     = {{RMR}: Re-consolidative Metacognitive Architecture for Identity Persistence in {LLM}},
  year      = {2026},
  publisher = {Zenodo},
  doi       = {10.5281/zenodo.19098594},
  note      = {Preprint}
}

@misc{Grosso2026cdos,
  author    = {Grosso, Daniele},
  title     = {{CDOS}: Cost-Driven Output Substitution---A Unified Framework for Sub-Threshold Behavioral Drift and Sycophancy in {LLM}s},
  year      = {2026},
  publisher = {Zenodo},
  doi       = {10.5281/zenodo.19397835},
  note      = {Preprint}
}

@book{hutchins1995,
  author    = {Hutchins, E.},
  title     = {Cognition in the Wild},
  publisher = {MIT Press},
  address   = {Cambridge, MA, USA},
  year      = {1995}
}

@book{jurafsky2009speech,
  title={Speech and Language Processing: An Introduction to Natural Language Processing, Computational Linguistics, and Speech Recognition},
  author={Jurafsky, D. and Martin, J.H.},
  isbn={9780131873216},
  lccn={2008010335},
  series={Prentice Hall series in artificial intelligence},
  year={2009},
  publisher={Pearson Prentice Hall}
}

@article{Lane2015MemoryRE,
    author={Richard D. Lane and Lee Ryan and Lynn Nadel and Leslie S. Greenberg},title={Memory reconsolidation, emotional arousal, and the process of change in psychotherapy: New insights from brain science.},
    journal={The Behavioral and brain sciences},
    year={2015},
    volume={38},
    pages={e1},
    url={https://api.semanticscholar.org/CorpusID:11491141}
}

@article{liu2022pretrain,
  author    = {Pengfei Liu and Weizhe Yuan and et al.},
  title     = {Pre-train, Prompt, and Predict: A Systematic Survey of Prompting Methods in Natural Language Processing},
  journal   = {ACM Computing Surveys},
  volume    = {55},
  number    = {9},
  year      = {2022},
  doi       = {10.1145/3560815}
}

@article{nader2000, 
    author={Nader, Karim and Schafe, Glenn E. and LeDoux, Joseph E.},
    title={Fear memories require protein synthesis in the amygdala for reconsolidation after retrieval},  
    journal={Nature}, 
    volume={406}, 
    number={6797}, 
    pages={722--726}, 
    year={2000}, 
    doi={10.1038/35021052} 
}

@book{norman1993things,
  title={Things that Make Us Smart: Defending Human Attributes in the Age of the Machine},
  author={Norman, D.A.},
  number={no. 842},
  isbn={9780201581294},
  lccn={92036016},
  series={A William Patrick book},
  year={1993},
  publisher={Addison-Wesley Publishing Company}
}

@article{panadero2017,
  author  = {Panadero, E.},
  title   = {A Review of Self-Regulated Learning: Six Models and Four Directions for Research},
  journal = {Frontiers in Psychology},
  year    = {2017},
  volume  = {8},
  pages   = {422},
  doi     = {10.3389/fpsyg.2017.00422}
}

@inproceedings{reeves2025prompts,
  author    = {Brent Reeves and James Prather and et al.},
  title     = {Prompts First, Precision Later: Reviving the Vision of Natural Language Programming for Computing Education},
  booktitle = {Proceedings of the 25th Koli Calling International Conference on Computing Education Research},
  year      = {2025},
  month     = nov,
  doi       = {10.1145/3769994.3770039}
}

@article{schraw1994,
  author  = {Schraw, G. and Dennison, R.S.},
  title   = {Assessing metacognitive awareness},
  journal = {Contemporary Educational Psychology},
  volume  = {19(4)},
  pages   = {460--475},
  year    = {1994},
  doi = {https://doi.org/10.1006/ceps.1994.1033}
}

@book{vygotsky2012thought,
  title={Thought and Language, revised and expanded edition},
  author={Vygotsky, L.S. and Kozulin, A.},
  isbn={9780262517713},
  lccn={2011048675},
  series={The MIT Press Paperback Series},
  year={2012},
  publisher={MIT Press}
}

@article{wei2022,
  title={Chain-of-Thought Prompting Elicits Reasoning in Large Language Models},
  author={Wei, Jason and others},
  journal={arXiv preprint arXiv:2201.11903},
  year={2022}
}

@book{winograd1972understanding,
  title={Understanding Natural Language},
  author={Winograd, T.},
  isbn={9780127597508},
  lccn={77190108},
  year={1972},
  publisher={Academic Press}
}

@book{winograd1987understanding,
  title={Understanding Computers and Cognition: A New Foundation for Design},
  author={Winograd, T. and Flores, F.},
  isbn={9780201112979},
  lccn={85003856},
  series={Language and being},
  year={1987},
  publisher={Addison-Wesley}
}

\end{document}